\documentclass{article}

\usepackage[preprint]{neurips_2020}

\usepackage[utf8]{inputenc} 
\usepackage[T1]{fontenc}    
\usepackage{hyperref}       
\usepackage{url}            
\usepackage{booktabs}       
\usepackage{amsfonts}       
\usepackage{nicefrac}       
\usepackage{microtype}      
\usepackage{amsmath}
\usepackage{mathtools}
\usepackage{graphicx}
\usepackage{caption}
\usepackage[singlelinecheck=off]{subcaption}

\graphicspath{{./figures/}}

\newcommand{\boldx}{\mathbf{x}}
\newcommand{\boldh}{\mathbf{h}}
\newcommand{\bolde}{\mathbf{e}}
\newcommand{\boldW}{\mathbf{W}}
\newcommand{\relu}{\mathrm{ReLU}}
\newcommand{\ind}[1]{\mathbf{1}\{#1\}}

\title{Lymph Node Graph Neural Networks for Cancer Metastasis Prediction}

\author{%
  Michal Kazmierski and Benjamin Haibe-Kains\\
  Department of Medical Biophysics, University of Toronto \&\\
  Princess Margaret Cancer Centre\\
  Toronto, ON, Canada\\
  \texttt{michal.kazmierski at mail.utoronto.ca, benjamin.haibe.kains at utoronto.ca} \\
}

\begin{document}

\maketitle

\begin{abstract}
Predicting outcomes, such as survival or metastasis for individual cancer patients is a crucial component of precision oncology. Machine learning (ML) offers a promising way to exploit rich multi-modal data, including clinical information and imaging to learn predictors of disease trajectory and help inform clinical decision making. In this paper, we present a novel graph-based approach to incorporate imaging characteristics of existing cancer spread to local lymph nodes (LNs) as well as their connectivity patterns in a prognostic ML model. We trained an edge-gated Graph Convolutional Network (Gated-GCN) to accurately predict the risk of distant metastasis (DM) by propagating information across the LN graph with the aid of soft edge attention mechanism. In a cohort of 1570 head and neck cancer patients, the Gated-GCN achieves AUROC of 0.757 for 2-year DM classification and \(C\)-index of 0.725 for lifetime DM risk prediction, outperforming current prognostic factors as well as previous approaches based on aggregated LN features. We also explored the importance of graph structure and individual lymph nodes through ablation experiments and interpretability studies, highlighting the importance of considering individual LN characteristics as well as the relationships between regions of cancer spread.
\end{abstract}

\section{Introduction}

Accurate prediction of disease trajectory and treatment outcomes is a key element of precision oncology. Nowadays, clinicians have access to an unprecedented amount of data about individual patient, from basic health information to high-resolution medical imaging and molecular biomarkers. Integrating the various signals to extract actionable predictions, however, is becoming increasingly challenging for a human. The growing amount and variety of data generated in complex domains like oncology makes it increasingly difficult for clinicians to integrating these multi-modal data in their decision process. Machine learning (ML) offers a way to make use of all the available data, automatically learning the underlying complex disease patterns from large multi-modal datasets. By delivering accurate, individualised predictions of outcomes and possible course of disease, ML can help doctors make better, data-driven decisions, ultimately translating into a benefit for the patient. 

\begin{figure}[h!]
    \centering
    \includegraphics[width=.6\textwidth]{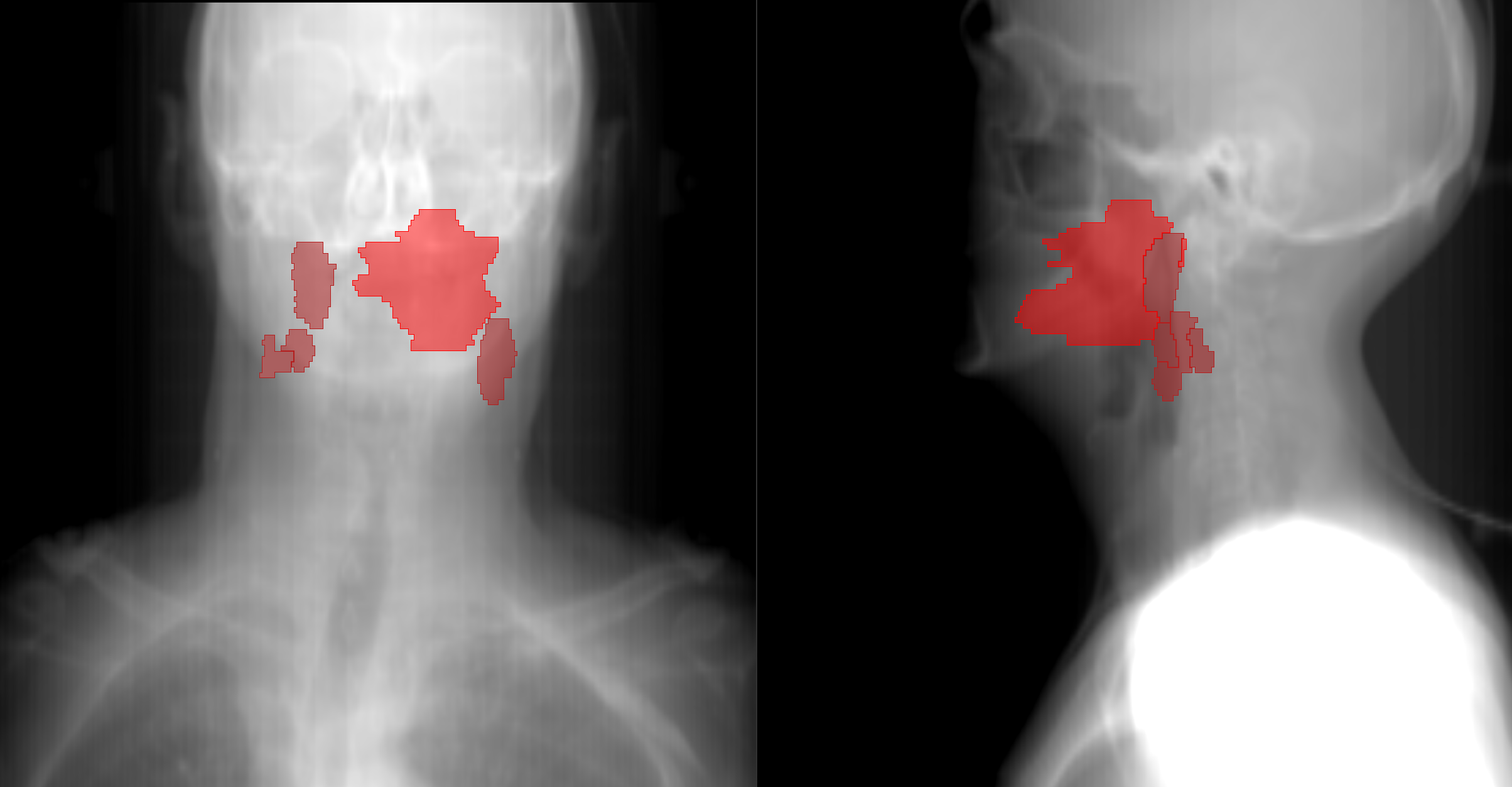}
    \caption{Coronal (left) and saggital (right) projections obtained from a CT scan of a HNC patient showing the primary tumour (lighter red near the centre) and invaded neck lymph nodes (darker red).}
    \label{fig:nodes}
\end{figure}

The need for new prognostic tools to guide clinical decision making is particularly apparent in head and neck cancer (HNC), with incidence of 900,000 cases worldwide every year that is projected to increase by 30\% by 2030 \citep{johnson_head_2020}. Despite recent advances in treatment, survival rates still remain suboptimal, largely due to the high heterogeneity in tumour biology and outcomes, making optimal treatment selection challenging. In particular, current prognostic factors are not sufficient to stratify patients by risk of distant metastasis (DM) --- the spread of cancer from the original site to other organs (such as lungs or brain) --- which is an indicator of particularly poor prognosis \citep{osullivan_deintensification_2013}. Local spread to lymph nodes (LNs) in the neck occurs frequently in HNC and often precedes distant metastasis by creating repositories of cancer cells with access to the lymphatic system \citep{lee_patterns_2018} \ref{fig:nodes}, but its current use in prognosis is limited to an aggregate score relying only on simple anatomical and imaging features.

We propose a novel machine learning approach to integrate tumour and LN imaging characteristics and clinical metadata to predict distant metastasis in head and neck cancer. We use a graph-based framework, with vertices representing metastatic LNs and edges the lymphatic connections in an individual patient-level graph. We leverage recent advances in geometric deep learning to develop an edge-gated Graph Convolutional Network (Gated-GCN) which learns the relationships between metastatic LNs in addition to their individual features and multi-task learning to accurately predict distant metastasis-free survival. We investigate the importance of graph structure and imaging features in ablation experiments and compare our approach to previously known prognostic factors.

\subsection{Related work}
\citet{vallieres_radiomics_2017} first proposed using computed tomography (CT) imaging characteristics of the primary tumour to predict distant metastasis in HNC, with encouraging results. Later, \citet{bogowicz_combined_2019} and \citet{wu_integrating_2019} introduced the idea of incorporating metastatic LN characteristics in predicting cancer recurrence and metastasis, respectively. The main limitation of these approaches is the reliance on aggregated features over all metastatic lymph nodes, ignoring the relationships between metastatic LNs as well as their potentially variable individual importance for prognosis. Apart from prognosis, \citet{chao_lymph_2020} used a graph neural network to reduce the false positive rate in mediastinal LN detection.

\section{Dataset}
 We used the RADCURE dataset \citep{kazmierski_machine_2021}, which includes pre-treatment CT scans and additional clinical metadata of HNC patients treated with radio(chemo)therapy at a single institution. Primary tumours and metastatic LNs were manually delineated by a trained radiation oncologist as part of the treatment planning process. We included patients with positive lymph node stage according to the TNM criteria \citep{brierley_tnm_2017} and where at least one LN contour could be retrieved, resulting in 1570 patients with median of 4 LN per patient (range 1--30). There were 259 DM events (16\%), with a median time to event of 1.1 years. 
 
 From each tumour and LN region of interest (ROI), we extracted a set of 11 image features previously identified as prognostic \citep{wu_integrating_2019}, as well as the coordinates of the region centroid, signed distance to the body mid-line and LN level (if applicable), which describes the position of a node in the lymphatic system of the neck \citep{gregoire_delineation_2014}. We also used the available standard prognostic factors, such as demographic information and staging.

\section{Graph-based encoding of lymph node relationships}
For each patient, we encoded the relationships between primary tumour and invaded LNs as an undirected graph \(G=(V, E)\). Each vertex, representing the tumour or LN, is associated with a vector of imaging features \(\boldx_i\), that is \(V = \{\boldx_i\}_{i=0}^{N_V}\) for an image with \(N_V-1\) metastatic LNs and one primary tumour. The edges represent the underlying connectivity pattern between the tumour and LNs and are initialized with the distance between the 3D centroids \(\mathbf{p}\) of the connected regions, \(e_{ij} = \Vert \mathbf{p}_i - \mathbf{p}_j \Vert\) for all \(e_{ij} \in E\). We note, however, that the graph convolutional network we used learns a more general edge update than just the distance which depends on the features of the connected vertices as well as previous edge features.

\subsection{Edge-gated GCN}
To learn prognostic representations from the input graphs we used the edge-gated graph convolutional network (Gated-GCN) \citep{bresson_residual_2018}. Gated-GCN learns anisotropic graph convolution operators by explicitly maintaining edge features at each layer and using them to compute dense soft attention coefficients which modulate vertex feature updates. Specifically, the vertex features at \((l+1)\)th layer are computed as
\begin{align}
    \boldh_i^{l+1} = \boldh_i^l + \relu(\boldW_1^l\boldh_i^l + \sum_{j \in \mathcal{N}_i}\boldsymbol{\alpha}_{ij}^l \odot \boldW_2^l\boldh_j^l),
\end{align}
where
\begin{align}
    \boldsymbol{\alpha}_{ij}^l = \frac{\sigma(\bolde_{ij}^l)}{\sum_{k \in \mathcal{N}_i}\sigma(\bolde_{ik}^l)},\quad  \sigma(z) = (1+\exp(-z))^{-1}
\end{align}
are the edge attention coefficients, \(\boldW_{1:5} \in \mathbb{R}^{d\times d}\) are learnable weight matrices for the hidden dimension \(d\) and \(\odot\) denotes elementwise product. The \(d\)-dimensional edge features are updated using
\begin{align}
    \bolde_{ij}^{l+1} = \bolde_{ij}^l + \relu(\boldW_3^l\bolde_{ij}^l + \boldW_4^l\boldh_i^l + \boldW_5^l\boldh_j^l).
\end{align}
Since the input vertex and edge feature dimensions do not necessarily match, we embed both in \(\mathcal{R}^d\) using a linear projection before the first graph convolutional layer. We also use batch normalization \citep{ioffe_batch_2015} on both node and edge features, as well as graph size normalization \citep{dwivedi_benchmarking_2020} to reduce the impact of variable input graph sizes. To help mitigate overfitting, dropout was applied to both the vertex and edge updates, as well as in the final graph-level fully-connected layers.

The graph structure and learned edge features enable the network to exploit the relationships between the existing regions of tumour spread in addition to their individual characteristics, which we hypothesise is crucial for accurate prediction of future cancer spread.

\subsection{Predictions and loss function}
\begin{figure}[t]
    \centering
    \includegraphics[width=\textwidth]{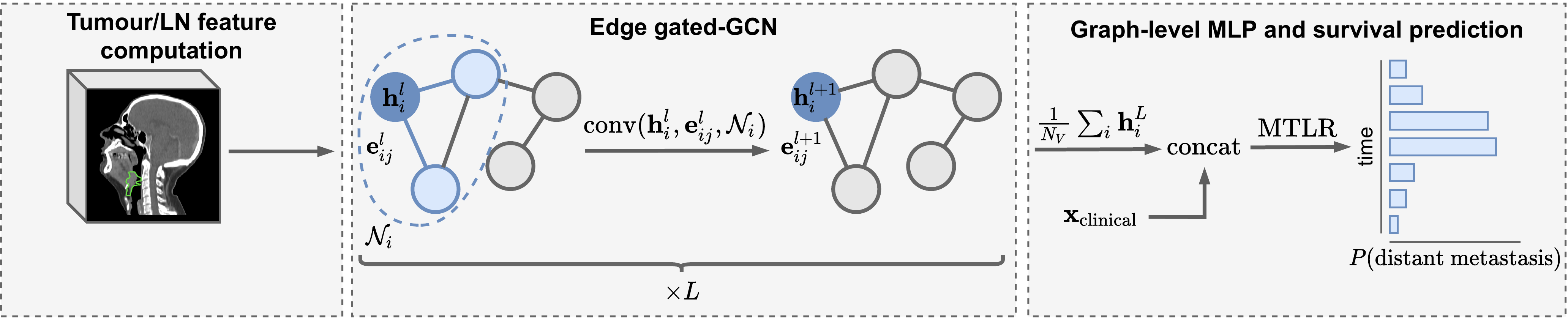}
    \caption{Overview of the proposed graph-based framework for DM prediction. Vertex features learned by \(L\) Gated-GCN layers (\(\mathrm{conv}\)) are pooled and concatenated with additional clinical metadata, \(\boldx_{\mathrm{clinical}}\). A graph-level MLP with an MTLR head is used to predict probability of DM jointly at multiple discrete timepoints.}
    \label{fig:graph_net}
\end{figure}

To predict the probability of distant metastasis over time, we used the multi-task logistic regression framework \citep{yu_learning_2011}. By predicting the probability of DM occurrence across multiple discrete time intervals in a multi-task way, it can better exploit the survival-type training data (also for patients with censored, i.e. partially observed survival times) and learn a flexible, time-varying risk function. Formally, we divide the time axis into \(K\) consecutive intervals \([t_{k-1}, t_k)\) for \(k=1, \dots, K\) and define a sequence of binary targets \(y_k = \ind{t_{k-1} \le T < t_k}\), where \(T\) denotes the time of DM occurrence and \(\ind{}\) is the indicator function. MTLR uses a separate set of parameters \(\{\mathbf{w}_k, b_k\}\) for each time interval, giving the predicted logits for event in a given interval as \(\hat{y}_k^{(j)} = (\mathbf{w}_k^T \mathbf{z}^{(j)} + b_k)\). The model is trained end-to-end using gradient descent by optimizing the following objective:
\begin{small}
\begin{align}
L(\mathbf{y}, \hat{\mathbf{y}}) &= -\sum_{j\ \mathrm{uncensored}}\sum_{k=1}^{K-1}y_k^{(j)}\times \hat{y}_k^{(j)} - \sum_{\mathclap{j\ \mathrm{censored}}}\log(\sum_{i=1}^{K-1}\ind{t_i \ge T_c^{(j)}}\exp(\sum_{k=i}^{K-1}y_k^{(j)}\times \hat{y}_k^{(j)})) + \sum_j Z^{(j)},
\label{eq:mtlr_likelihood}
\end{align}
\end{small}
where \(T_c^{(j)}\) denotes the censoring time of the \(j\)th patient and \(Z^{(j)} = \log(\sum_{i=1}^{K}\exp(\sum_{k=i}^{K-1}\hat{y}_k^{(j)}))\) is a normalizing constant.
In the Gated-GCN case, we set \(\mathbf{z} = \mathrm{MLP}(\boldh_{\mathrm{out}})\), where \(\boldh_{\mathrm{out}} = \frac{1}{N_V}\sum_{i=1}^{N_V}\boldh_i^{L}\) are the pooled vertex features from the final graph convolutional layer, potentially concatenated with the additional patient-level clinical covariates and \(\mathrm{MLP}\) is a graph-level multi-layer perceptron (fig. \ref{fig:graph_net}).

We derived two kinds of predictions from the MTLR output: the probability of event before a specified time point (here 2 years) and the lifetime risk of DM, which takes into account the predictions at all timepoints.

\section{Experiments}
We evaluated the performance of our approach on 2 tasks: predicting distant metastasis before 2 years (a binary classification task, with target \(y_{\mathrm{bin}} = \ind{\text{patient develops DM before 2 years post-treatment}}\)) using area under the ROC curve (AUROC) as the performance metric and predicting the lifetime metastasis risk (a survival prediction task), using the concordance (\(C\)) index \citep{harrell_multivariable_1996}. The inclusion of the binary classification task enabled us to evaluate the performance of the proposed approach at identifying high-risk patients that might require additional interventions.

We experimented with 2 types of graph connectivity: (1) complete (with edges between all pairs of vertices) and (2) \(K\)-nearest neighbours (where every vertex is connected to \(K\) closest vertices by Euclidean distance) for \(K=2,\ 3\).

To investigate the role of graph structure, we trained a graph-agnostic neural network which shares weights across the vertices, similarly to the Gated-GCN, but ablates the graph connectivity \citep{luzhnica_graph_2019}. The graph-level feature vector is obtained by mean-pooling the final layer features across the vertices. We also compared the performance of the proposed approach to a set of simple benchmark models, which use known prognostic clinical features alone or combined with either aggregated tumour and LN features, tumour features only or total tumour + LN volume. We used the neural MTLR architecture \citep{fotso_deep_2018} with one or more hidden layers for all of the simple baselines.

In all experiments, we used the same 5-fold stratified cross-validation training setup. We trained each model for 100 epochs using the Adam optimizer \citep{kingma_adam:_2014}. Hyperparameters, including the learning rate, batch size, number and width of hidden layers and dropout rate were tuned using nested random search. The proposed approach and all benchmark models were implemented using PyTorch \citep{paszke_pytorch_2019}, Lightning \citep{falcon_pytorch_2019} \texttt{pytorch-geometric} \citep{fey_fast_2019} and \texttt{torchmtlr}\footnote{\url{https://github.com/mkazmier/torchmtlr}}. Details of the training and evaluation protocol can be found in Appendix \ref{appendix:training}.



\section{Results}
The best performance in both 2-year and lifetime risk prediction was achieved by the GCN trained on fully-connected graphs (table \ref{tab:results}). All models using imaging features achieved better prognostic performance than baseline relying on clinical metadata only, with further increase in performance when using LN features in addition to the primary tumour. Notably, both the GNN and graph-agnostic NN performed better than the clinical-only baseline, indicating that LN imaging features can be strong, independent predictors of distant metastasis. Furthermore, considering the LN characteristics separately was necessary for higher predictive accuracy (which is further supported by the variable importance of different metastatic LNs in the model, as described in the Interpretability section below).

\begin{table}[h!]
\centering
\caption{Performance of the Gated-GCN with varying graph connectivity, graph-agnostic neural networks and simple baselines on the classification and lifetime risk tasks. Metrics are reported as mean and standard deviation over the 5 folds.}
\begin{tabular}{lrrrrrr}
    \toprule
    {} & \multicolumn{2}{c}{\textbf{2-year AUROC}} & \multicolumn{2}{c}{\(C\)\textbf{-index}} \\
    \cmidrule(lr){2-3} \cmidrule(lr){4-5}
    {} &      mean &   std &  mean &   std \\
    \midrule
    \textbf{Gated-GCN + clinical (complete graph)} &  \textbf{0.757} & \textbf{0.058} & \textbf{0.725} & \textbf{0.016} \\
    Gated-GCN + clinical (\(K=3\))                 &     0.750 & 0.026 & 0.719 & 0.021 \\
    Gated-GCN + clinical (\(K=2\))                 &     0.747 & 0.021 & 0.717 & 0.024 \\
    \\
    Graph-agnostic NN + clinical             &     0.726 & 0.029 & 0.696 & 0.017 \\
    \midrule
    Clinical + tumour/node features          &     0.708 & 0.039 & 0.683 & 0.024 \\
    Clinical + tumour/node volume            &     0.693 & 0.053 & 0.665 & 0.035 \\
    Clinical + tumour features only          &     0.682 & 0.067 & 0.653 & 0.046 \\
    \\
    Clinical only                            &     0.668 & 0.053 & 0.648 & 0.037 \\
    \bottomrule
\end{tabular}
\label{tab:results}
\end{table}

\subsection{Role of graph structure}
The best graph-based model achieved higher AUROC (.757) and \(C\)-index (.725) than the graph-agnostic baseline (\(\mathrm{AUROC}=.726,\ C=.696\)),
indicating that explicitly modelling relationships between the existing metastases is important for predicting further cancer spread. The performance was similar across different choices of graph topology, with slight bias towards stronger connectivity. This suggests that the network might be able to make use of greater number of connections per vertex, with the edge attention process modulating the importance of neighbours during updates (with 3-nearest neighbour connectivity many of the smaller graphs become complete).

\subsection{Interpretability}
\begin{figure}[h!]
    \centering
    \begin{subfigure}[b]{.55\textwidth}
        \caption{}
        \label{fig:graph1}
    \end{subfigure}
    \begin{subfigure}[b]{.35\textwidth}
        \caption{}
        \label{fig:igs1}
    \end{subfigure}
    \vfill
    \includegraphics[width=\textwidth]{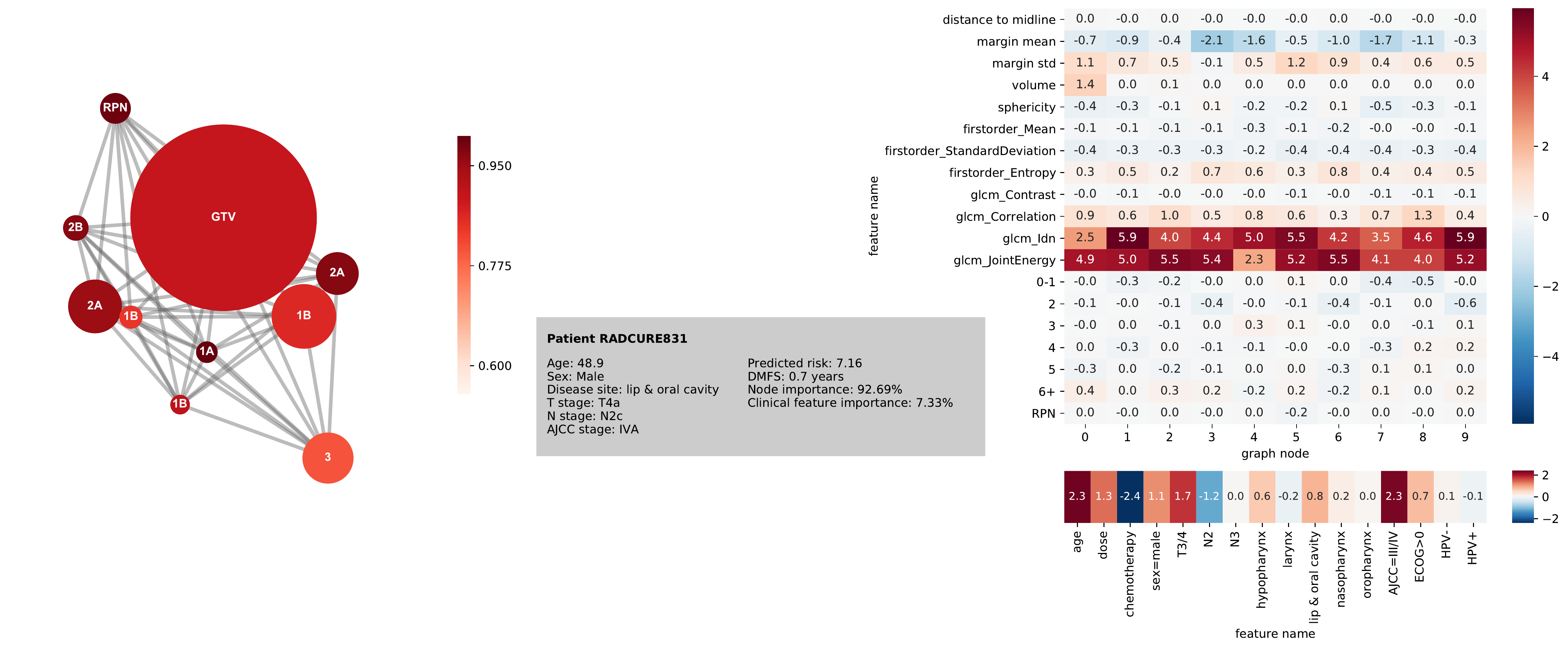}
    \begin{subfigure}[b]{.55\textwidth}
        \caption{}
        \label{fig:graph2}
    \end{subfigure}
    \begin{subfigure}[b]{.35\textwidth}
        \caption{}
        \label{fig:igs2}
    \end{subfigure}
    \includegraphics[width=\textwidth]{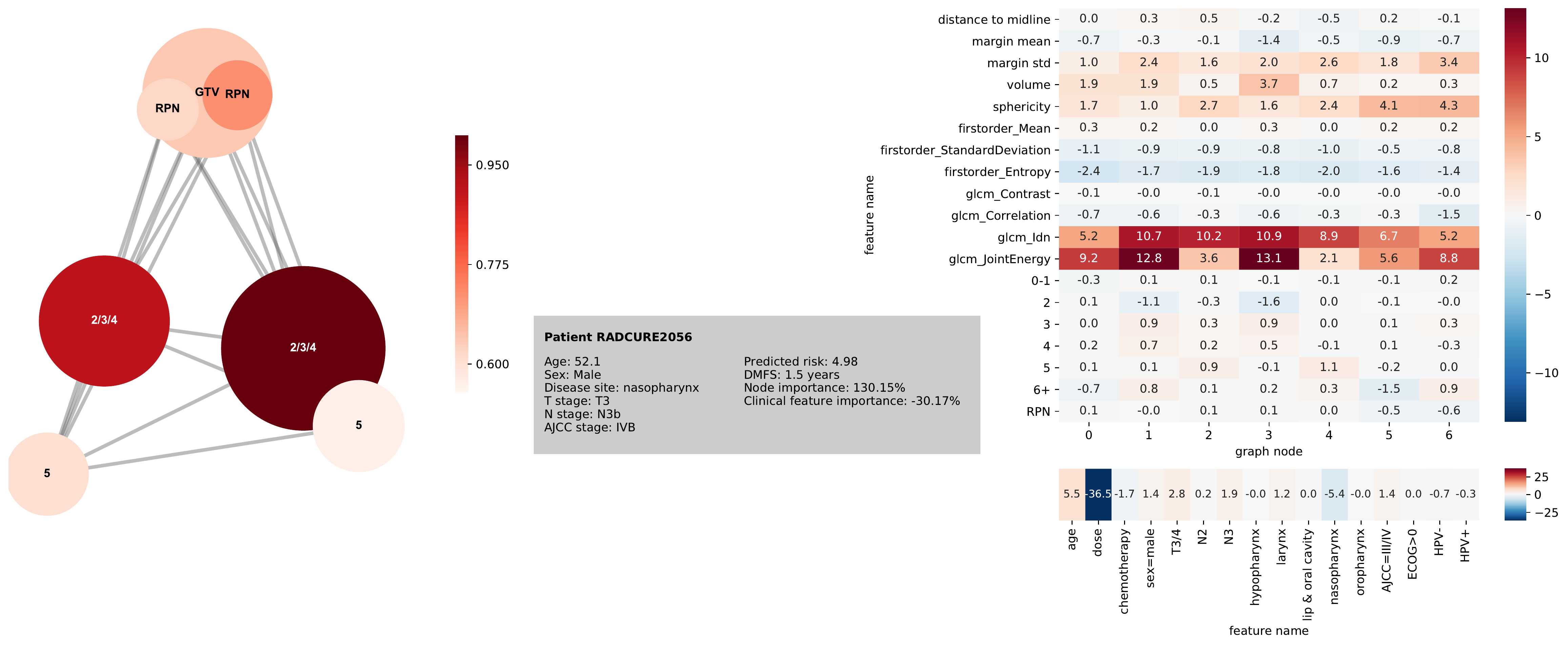}
    \caption{Attributions for randomly selected patients with time to DM shorter (top) and longer (bottom) than median. \ref{fig:graph1} and \ref{fig:graph2} show the input graph, where the size of each node is proportional to the volume of the associated region and the colour intensity corresponds to the total attribution score normalized by the maximum score. The vertices are labeled by the nodal level (number or \texttt{RPN} for retropharyngeal nodes) or \texttt{GTV} for the primary tumour. \ref{fig:igs1} and \ref{fig:igs2} show the normalized integrated gradients for node features (top) and clinical metadata (bottom).}
    \label{fig:int_grads}
\end{figure}

Understating the factors contributing to the model's prediction can be just as important as high discriminative performance for both doctors and patients, particularly when the predictions factor into clinical decisions \citep{tonekaboni_what_2019}. We used integrated gradients \citep{sundararajan_axiomatic_2017} to evaluate the contributions of the graph vertices, as well as individual vertex and patient-level features to the predictions of the best model (fig. \ref{fig:int_grads}). Vertex-level attribution scores were computed as sums of absolute values of feature attributions for each vertex, normalized by sum of attributions over all vertices.

The attributions can be used to develop a qualitative intuition of how the model operates. Inspecting vertex integrated gradients for a random sample of test patients revealed variable relevance of individual vertices for the predictions. Notably, in some cases a LN received over 60\% higher attributions than the primary tumour. Additionally, we found that several features had consistently high importance for model predictions, including 2 image heterogeneity features and a region margin measure.

\section{Conclusions}
We presented a method for graph-based distant metastasis prediction in head and neck cancer. Our approach using edge-gated graph convolutional network achieved better performance than the graph-agnostic baselines on both 2-year classification and lifetime risk prediction, and is a promising new way of incorporating data from multiple regions of interests and their relationships in a prognostic model. Additionally, our results highlight the relevance of lymph node imaging biomarkers for DM prediction. Although we used HNC as a case study, due to high incidence of LN metastases, our approach is straightforward to apply in other cancer sites where local LN spread commonly occurs, e.g. breast.

The study has several potential limitations. The dataset we used was relatively small (although still substantial in comparison with other publicly-available HNC datasets) and single-institution. We plan to extend our dataset with more patients, as well as test the proposed approach using data from a different hospital. Additionally, our model relies on engineered image features. It is likely that learned representations from a convolutional network could further improve the performance of our method and we are working on an end-to-end approach learning directly from images. We leave this for future work.

\medskip

\small

\bibliographystyle{agsm}
\bibliography{references}

\appendix
\section{Dataset details}
\label{appendix:data}

The detailed description of the clinical and imaging data can be found in \citet{kazmierski_machine_2021}. Here we highlight the some of the details relevant to the present study.

We used the following clinical metadata: age, sex, tumour location and stage, overall health status (ECOG criteria), radiation dose, chemotherapy use and human papillomavirus (HPV) infection status. Continuous features were standardized to zero mean, unit variance and discrete features were one-hot encoded, with missing values handled by adding additional missingness indicator level.

CT features were computed using PyRadiomics version 3.0 \citep{van_griethuysen_computational_2017}, with the exception of margin features, for which we used an in-house implementation following \citep{levman_margin_2011}. We used the same feature set and preprocessing protocol as \citet{wu_integrating_2019}. The following image features were extracted from each region of interest: margin mean and standard deviation, volume, sphericity, firstorder-Mean, firstorder-StandardDeviation, firstorder-Entropy, GLCM-Contrast, GLCM-Correlation, GLCM-Idn (equivalent to GLCM-Homogeneity1), GLCM-JointEnergy, distance to midline and lymph node level. For the lymph node benchmark model, the features were averaged over all lymph nodes (except volume which was summmed) and additional features were added (LN count, maximum tumour-node and node-node distance), following \citep{wu_integrating_2019}.

\section{Training and hyperparameter selection}
\label{appendix:training}
 For each cross-validation split, we initially reserved one of the training folds for validation and tuned the hyperparameters using random search (see table \ref{tab:hyperparams} for hyperparameter distributions). We selected the hyperparameter set with the lowest validation loss out of 60 search iterations and re-trained the model on all training folds before evaluating on the test fold.
 
 \begin{table}[h!]
  \centering
    \caption{Hyperparameter distributions used in random search. Square brackets indicate discrete uniform distribution over the values. \(\mathrm{loguniform}(a, b)\) is the uniform distribution in log domain between \(\log(a)\) and \(\log(b)\).}
  \begin{tabular}{ll}
    \toprule
    Hyperparameter & Distribution\\
    \midrule
    Learning rate & \(\mathrm{loguniform}(10^{-4},\ 5\times10^{-3})\)\\
    Batch size & \([64, 128, 256, 512]\)\\
    Weight decay & \(\mathrm{loguniform}(10^{-5},\ 10^{-3})\)\\
    MTLR \(\ell_2\) regularization & \([10^{-2}, 10^{-1}, 10^{0}, 10^{1}, 10^{2}, 10^{3}]\)\\
    Dropout probability & \(\mathrm{uniform}(0,\ .5)\)\\
    Number of GCN/node-level MLP layers & \([1, 2, 3, 4]\)\\
    Number of MLP layers & \([1, 2, 3, 4]\)\\
    Layer width & \([32, 64, 128, 256]\)\\
    \bottomrule
  \end{tabular}
  \label{tab:hyperparams}
\end{table}

\end{document}